\documentclass[preprint,11pt,authoryear]{elsarticle}

\usepackage{amsmath}
\usepackage{amsfonts}
\usepackage{amssymb}

\usepackage{float}
\usepackage{bm}
\usepackage{caption}
\usepackage{subcaption}
\usepackage{lettrine}
\usepackage{rotating}
\usepackage{lscape}
\usepackage{threeparttable}
\usepackage{mathtools}
\usepackage[table]{xcolor}
\usepackage{wasysym}

\usepackage[english]{babel}
\usepackage{booktabs}
\usepackage{cases}
\usepackage{url}

\journal{European Journal of Operational Research}

\begin{document}

\begin{frontmatter}



\title{Optimization with Constraint Learning: A Framework and Survey}

\author{Adejuyigbe O. Fajemisin}
\ead{a.o.fajemisin2@uva.nl}
\author{Donato Maragno}
\ead{d.maragno@uva.nl}
\author{Dick den Hertog}
\ead{d.denhertog@uva.nl}

\address{Amsterdam Business School, University of Amsterdam, 1018TV Amsterdam,The Netherlands}

\begin{abstract}
Many real-life optimization problems frequently contain one or more constraints or objectives for which there are no explicit formulae. If however data on feasible and/or infeasible states are available, these data can be used to learn the constraints. The benefits of this approach are clearly seen, however, there is a need for this process to be carried out in a structured manner. This paper, therefore, provides a framework for Optimization with Constraint Learning (OCL) which we believe will help to formalize and direct the process of learning constraints from data. This framework includes the following steps: (i) setup of the conceptual optimization model, (ii) data gathering and preprocessing, (iii) selection and training of predictive models, (iv) resolution of the optimization model, and (v) verification and improvement of the optimization model. We then review the recent OCL literature in light of this framework and highlight current trends, as well as areas for future research.
\end{abstract}

\begin{highlights}
\item We formalize the process of optimization with constraint learning.
\item We review the literature using constraint learning in light of the formalization.
\item Current trends are discussed and opportunities for future work are highlighted.
\end{highlights}


\end{frontmatter}


    \begingroup
    
      \renewcommand{\cleardoublepage}{} 
      \renewcommand{\clearpage}{} 
      
      \newpage
\section{Introduction} 
\label{sec:intro}
\subsection{The Synergy between Optimization and Machine Learning}
\label{sec:synergy}
The synergy between mathematical optimization and Machine Learning (ML) has come to the forefront in recent years and has been discussed in depth in the recent literature. \cite{sun2019survey} provide a recent survey of optimization methods from an ML perspective. In another survey, \cite{gambella2021optimization} present fundamental ML tasks such as classification, regression, clustering, adversarial learning, and others as optimization problems. In the optimization field, more and more attention is being given to using ML to aid the formulation and solution of optimization problems. Practical optimization problems can be large and difficult to solve, and more and more authors are exploiting the flexibility and usefulness of ML to aid in the solution of difficult optimization problems (e.g. \cite{FISCHETTI2019} and \cite{abbasi2020predicting}). In addition to speeding up the solution of problems, ML and deep learning have also been used to speed up tree search in \cite{HOTTUNG2020}, accelerate the Branch-and-Price Algorithm in \cite{VACLAVIK2018}, and guide branching in Mixed Integer Programs (MIP) in \cite{KHALIL2016}. ML has also been used to improve meta-heuristic solution approaches \citep{karimi2022machine}. A recently published survey by \cite{BENGIO2020} provides an overview of recent attempts to leverage machine learning to solve combinatorial optimization problems. 

\subsection{Motivation for this Study}
\label{sec:motiveForStudy}
The examples in the previous section use ML to learn or obtain solutions to optimization problems in a more efficient manner. In our view, however, there is even more value in using ML for optimization: ML can be used to learn constraints. Many real-life optimization problems contain one or more constraints or objectives for which there are no explicit formulae. However when there is data available, one can use the data to learn the constraints. Two such examples are given in Section \ref{sec:examples}. These examples show the difficulty of formulating some constraints using explicit formulae and highlight the usefulness of being able to learn such constraints from available data. In addition to using traditional ML approaches to learn constraints for optimization problems (e.g. regression trees and linear regression in \cite{verwer2017auction} and neural networks in \cite{villarrubia2018artificial}), other techniques such as genetic programming (e.g. in \cite{pawlak2018synthesis}), mixed-integer linear programming (MILP) (e.g. in \cite{pawlak2017automatic}) and symbolic regression have also been used. In our literature search, we have noticed several works of this kind from several different application areas and even several scientific fields. However, it seems beneficial to have these works collated and discussed in one place. Additionally, we propose to take a more holistic view of the entire constraint learning process, starting from defining the base problem, to the resolution and validation of the final optimization model.

This paper is therefore intended to provide both a survey of literature on constraint learning for optimization, as well as offer a framework for effectively facilitating Optimization with Constraint Learning (OCL). The five steps of the proposed framework are: (i) setup of the conceptual optimization model, (ii) data gathering and processing, (iii) selection and training of the predictive model(s), (iv) resolution of the optimization model, and (v) validation/improvement of the optimization model. These steps will be discussed in more detail throughout this paper. We use the term ``Constraint Learning" to make clear that we are interested in cases in which most of the optimization model is known, but some of the constraints are difficult to model explicitly. This is to differentiate from other terms such as ``model learning" \citep{donti2017task} and ``model seeking" \citep{beldiceanu2016modelseeker} seen in the literature. We should also note that there are several approaches in the literature that are similar to \textit{Constraint Learning} but do not fit in this survey for several reasons. In Constraint Programming (CP) for example, the process of learning CP models from data (positive and/or negative examples) has been widely studied \citep{bessiere2004leveraging,o2010automated,beldiceanu2016modelseeker,kolb2016learning,galassi2018model}. These are not included as we are interested in cases where predictive models are embedded as constraints in mathematical optimization. 

We also exclude papers on the subject of Satisfiability Modulo Theories (SMT) \citep{nieuwenhuis2006sat} from our survey. Additionally, papers such as \cite{bertsimas2020predictive}, \cite{elmachtoub2022smart}, \cite{demirovic2019predict+} and \cite{villarrubia2018artificial} which focus on predicting problem \textit{parameters}, or use predictive models to approximate \textit{known} objective functions and constraints are excluded. We also exclude papers where the structure of the problem is known, but predictive models are used to replace the problem in order to speed up computation \citep{nascimento2000neural,nagata2003optimization,venzke2020learning,venzke2020verification}. Similarly, papers like \cite{shang2017data} and \cite{han2020multiple} which use predictive models to learn the uncertainty set for robust optimization problems are not considered. Finally, papers which show an interplay between ML and optimization in some way but do not focus on constraint learning (e.g. \cite{gilan2015sustainable}, \cite{bagloee2018hybrid}, \cite{say2018planning} and \cite{say2020theoretical}) are also out of the scope of this survey.

While \cite{lombardi2018boosting} and \cite{kotary2021end} provide short surveys on OCL, we present a more extensive framework for constraint learning and consider a wider range of the literature. Papers identified in our literature search will be discussed in light of the proposed framework, and gaps in knowledge and scope for future lines of research will be identified. Running examples will be used to illustrate the steps of the framework. It is envisaged that in addition to providing a review as well as the framework, our work will bring together ideas from diverse fields.

\subsection{Outline}
\label{sec:introOutline}
This framework and survey is structured as follows. In Section \ref{sec:examples}, we first present the running examples that will be used to illustrate our framework. We then describe this framework in Section \ref{sec:genDecr} and review the recent literature on light of it in Section \ref{sec:review}. Areas for further research are presented in Section \ref{sec:challengesandOpp}, with general conclusions given in Section \ref{sec:Conclusions}. 
      \section{Running Examples}
\label{sec:examples}
In order to illustrate the concepts presented in later sections, we first present two practical optimization problems in which constraint learning plays an important role. 

\subsection{Palatability Considerations in WFP Food Baskets}
\label{sec:WFPExample}
This first example deals with the issue of providing palatable rations in the food baskets delivered by the World Food Programme (WFP) to population groups in need. These food baskets are designed to meet the nutritional requirements of a population and take into account existing levels of malnutrition and disease, climatic conditions, activity levels, and so on \citep{wfpwebsite}. In addition to considering the nutritional content, we would also like to be able to consider the palatability of the rations. This problem is particularly relevant in that the palatability of a food basket can change drastically between different groups of beneficiaries. Since the palatability constraint is difficult to model manually, and a data set with different food baskets and their palatability scores is available, a predictive model can be used to derive the palatability constraint from the data, which can then be embedded into the original food basket optimization problem. 

Let $\mathcal{K}$ be the set of commodities (such as rice, flour, chickpeas, sugar, etc.) that may be included in a food basket, and $\mathcal{L}$ be the set of nutrients (such as vitamins, minerals, proteins, etc.) which are important for maintaining good health. If the nutritional requirement for nutrient $\textit{l} \in \mathcal{L}$ (in grams/person/day) is denoted by $Nutreq_{l}$, and the nutritional value for nutrient \textit{l} (per grams of commodity $\textit{k} \in \mathcal{K}$) is denoted by $Nutval_{kl}$, then a simple model for food basket optimization problem can be given as
\begin{subequations}
\begin{align}
\min_{\bm{x}} & \ \bm{c}^T\bm{x} \label{eqn:rationCost}\\
\mbox{s.t.} \ & \sum_{k \in \mathcal{K}} Nutval_{kl} \, x_{k} \geq Nutreq_{l} & \forall l \in \mathcal{L} \label{eqn:nutrients}\\
& Palatability\_score(\bm{x}) \geq P_{min} \label{eqn:palatability}\\
& x_{k} \geq 0 & \forall k \in \mathcal{K}, \label{eqn:nonNeg}
\end{align}
\end{subequations}

\noindent
where $x_k$ is the amount of each commodity $k \in \mathcal{K}$ (measured in grams) included in the food ration, $\bm{c}\in \mathbb{R}^{|\mathcal{K}|}$ is the cost vector associated with the commodities (in \$/gram), and the objective function (\ref{eqn:rationCost}) minimizes the ration cost. Constraint (\ref{eqn:nutrients}) is used to ensure a minimum level of nutritional requirements in the rations. Constraint (\ref{eqn:palatability}) has been learned using a predictive model and is used to ensure that the palatability of the ration is at least equal to $P_{min}$. The diet problem in \eqref{eqn:rationCost} to \eqref{eqn:nonNeg} is only a part of a much bigger model (see \cite{peters2021nutritious}) that also includes the sourcing, facility location and transportation problems. The constraint to be learned here is a function of relatively few variables ($|\mathcal{K}|$ in this case), compared with the thousands of variables in the full model.

The predictive model is trained on a sparse data set $\bm{D}$ in $\mathbb{R}^{n\times(|\mathcal{K}|+1)}$ composed of $n$ samples, with each sample composed of $|\mathcal{K}|$ features and one label. Each sample is a different food basket where the $|\mathcal{K}|$ features are the amounts of each commodity $k$ that make up the food basket, and the label corresponds to the palatability score associated with each specific food basket. This label can either be numeric or categorical, resulting in $Palatability\_score(\bm{x})$ being either a regression or classification model respectively. The data and code for this example are available at \texttt{https://github.com/hwiberg/OptiCL}. 


\subsection{Radiotherapy Optimization}
\label{sec:RadOptExample}
The second example problem we will use to illustrate the framework is radiotherapy (RT) optimization. In RT, ionizing radiation beams are used to control tumor growth by damaging cancerous tissue. Healthy, non-cancerous tissue (also known as organs-at-risk (OAR)) are however present in close proximity to the tumors, and consequently incur unavoidable damage. The tumor damage can be quantified using the Tumor Control Probability (TCP), while the unavoidable damage to the OARs can be quantified using the Normal Tissue Complications Probability (NTCP). The final optimization problem is to maximize the TCP subject to a set of constraints ensuring that the NTCP does not exceed a specified level. If $\mathcal{Q}$ is the set of OARs (e.g. stomach, liver, spinal cord) that need to be protected, then the optimization problem can be written as
\begin{subequations}
\begin{align}
\max_{\bm{x}, \bm{d}} & \ \text{TCP}(\bm{d}, \bm{\gamma}) & \label{eqn:TCPobj}\\
\mbox{s.t.} \ & \text{NTCP}_q(\bm{d}, \bm{\gamma}) \leq \delta_q & \forall q \in \mathcal{Q} \label{eqn:NTCP}\\
& f(\bm{d}) \leq 0 & \label{eqn:doseRestr}\\
& \bm{\mathrm{A}} \bm{x} = \bm{d} & \label{eqn:dose}\\
& \bm{x} \geq 0, & \label{eqn:beamlet}
\end{align}
\end{subequations}

\noindent
where $\bm{\gamma}$ is a vector containing the biological characteristics of the tumor and the OARs, as well as other patient-specific information. The maximum allowable damage to OAR $q$ is given by $\delta_q$, while constraint \eqref{eqn:doseRestr} is a placeholder for other constraints on the dose distribution $\bm{d}$ (e.g. minimum dose, maximum dose, etc.) and is usually convex or can be converted to a convex form. The decision variables $\bm{x}$ and $\bm{d}$ are the set of the radiation intensities and the doses to be delivered over the course of the treatment respectively, while $\bm{\mathrm{A}}$ is the dose deposition matrix. A more detailed treatment of intensity-modulated radiation therapy optimization can be found in \cite{hoffmann2008convex}.  

In the current literature, explicit expressions are used for the TCP and NTCP functions \citep{GAY2007115}. These however are surrogate functions based on population-wide estimates, and as such, the response of different patients to the same treatment can be quite varied \citep{SCOTT2020}. Thus, while some patients experience complete remission with minimal damage to OARs, others suffer from radiation poisoning or tumor recurrence. It has been shown in \cite{TUCKER2013251} that the inclusion of patient-specific characteristics (such as genetic predispositions) can improve the prediction of patients' response to RT treatment. We therefore intend to use data to learn \eqref{eqn:TCPobj} and \eqref{eqn:NTCP}, leading to the creation of more effective treatment plans for patients.\\

The above two examples have illustrated the potential benefits of OCL. In the next section, we will describe the general framework for OCL, as well as show how to apply the steps of the proposed framework to optimization problems.

      \section{General Description of Framework}
\label{sec:genDecr}
The proposed framework is the result of a careful analysis of many approaches adopted in the literature covered by this survey. It is the formalization of a procedure that (partially) recurs in most approaches to learning constraints. We believe that a complete view of all the steps that characterize the framework will help future research see the big picture of constraint learning, and the impact that each step has on the final performance.

\subsection{Step 1: Setup of the Conceptual Optimization Model}
\label{sec:overviewStep1}
This initial step is usually done in collaboration with domain experts. This step begins with (i) defining the decision variables and parameters involved, and (ii) defining those constraints that are easy to model manually. After this, the constraint(s) to be learned can be included. Finally, any additional auxiliary constraints required to facilitate the constraint learning process may be added. A broad formulation of all the fundamental aspects that characterize optimization with constraint learning is given as:

\begin{subequations}
\begin{align} 
\min_{\bm{x},\bm{y}} \ & f(\bm{x}) & \label{eqn:step1Ebegin} \\
\mbox{s.t.} \ & \bm{g}(\bm{x}) \leq 0 & \label{eqn:step1Eknown}\\
& \bm{y} = \hat{\bm{h}}(\bm{x},\bm{w}, \bm{z}) & \label{eqn:step1Epred} \\
& z_i = s_i(\bm{x}) & i = 1, \ldots, m \label{eqn:step1Efeature}\\
& \bm{\theta}(\bm{y}) \leq 0 & \label{eqn: step1Econstr} \\
& \bm{x} \in X, & \label{eqn:step1Eend}
\end{align}
\end{subequations}

\noindent
where the vector $\bm{x}$ is the vector of decision variables, \eqref{eqn:step1Ebegin} is the known objective function and \eqref{eqn:step1Eknown} are the constraints that are easy to model. In \eqref{eqn:step1Epred}, $\bm{y}$ is the difficult part of the model to design manually and is the output of a predictive model $\hat{\bm{h}}(\mathord{\cdot})$ which takes as input decision variables $\bm{x}$, non-decision variables $\bm{w}$, and $\bm{z}$, which is the vector of variables designed as a combination of the $\bm{x}$ variables. 
The predictive model's performance can be often improved with the addition of other features $\bm{w}$ that are not part of the decision-making process, but convey additional or contextual information about the problem \citep{ChenSide2020,bertsimas2022dynamic}. For example in the radiotherapy optimization example in Section \ref{sec:RadOptExample}, although the decision variables are radiation intensities $\bm{x}$, additional information is conveyed by the vector $\bm{\gamma}$ which contains patient-specific information (such as gene expressions). Although $\bm{w}$ is known in this example, it may be unknown in other problems and as such, we may have to optimize for the average or worst case. It may also be possible to treat any uncertainties in $\bm{w}$ using a robust optimization \citep{gorissen2015practical} or stochastic programming \citep{birge2011introduction} approach.

The vector $\bm{x}$ can also be exploited to build more complex features using a process known as \textit{feature engineering} \citep{kuhn2019feature}. The idea behind this process is to use effective combinations of decision variables as new features to improve the performance of the predictive model, reduce the number of features used in the learning process, or embed prior knowledge into the predictive model. These new features $\bm{z} = (z_1,\ldots,z_m)$ can be expressed using constraint \eqref{eqn:step1Efeature}, where $s_i$ is the specific function used to extract the feature $z_i$. The palatability constraint in Section \ref{sec:WFPExample} is an example of such a case. It is not simply a function of the decision variables, but more accurately a function of aggregates of $\bm{x}$. These $z_i$ variables represent the amount of food per macro category (Cereals \& Grains, Pulses \& Vegetables, Oils \& Fats, Mixed \& Blended foods, Meat \& Fish \& Dairy), so that \eqref{eqn:palatability} becomes $Palatability\_score(\bm{z}) \geq P_{min}$. It should be noted that some part of the known constraint \eqref{eqn:step1Eknown} might also be a function of $\bm{y}$, i.e. $\bm{g}(\bm{x})$ could be $\bm{g}(\bm{x}, \bm{y})$, however we will continue with the definition where the known and unknown constraints are written separately. Similarly, some part of the objective function might also depend on the quantity to be learned (i.e. $f(\bm{x},\bm{y})$), but we will also continue with the definition where $f(\bm{x})$ is known. Finally \eqref{eqn: step1Econstr} contains any required constraints on $\bm{y}$. 
 
In order to avoid getting a solution far from what is seen in the data, we can restrict the optimization to a region that is densely populated by samples that have been used to train and test the predictive model. This region is known as a Trust Region (TR), and is briefly mentioned by \cite{biggs2018optimizing} and expounded upon in \cite{maragno2021mixedinteger}. Here, the authors constrain the optimal solution to be within the convex hull of all samples (training and test samples regardless of their label). Although the computation of the convex hull is burdensome in a high-dimensional space, the authors circumnavigate the problem by constraining the optimal solution to be a convex combination of the samples. Thus, the computation of the convex hull facets is unnecessary and the problem distills to constraining the optimal solution with $O(n)$ constraints, where $n$ is the dimensionality of the decision variable.

Ideally, a trust region should account for outliers and prevent low-density regions to be considered trustworthy. \cite{maragno2021mixedinteger} propose a two-step approach that first identifies high-density clusters and then defines the trust region as a combination of multiple convex hulls, one for each cluster. Smaller and denser trust regions lead to worse solutions in terms of objective function value, but also to greater confidence in the predictive model evaluation. A similar approach in terms of clustering high-density regions before constraint learning is seen in \cite{sroka2018one}, and \cite{karmelita2020cma}.

\subsection{Step 2: Data Gathering and Preprocessing}
\label{sec:overviewStep2}
Once the conceptual optimization model has been formulated, the next step is to focus on collecting and preprocessing the data which will be used to learn the constraints \eqref{eqn:step1Epred}. As data gathering and processing is a huge subject in its own right, we will only discuss those aspects that are important in the context of constraint learning. The characteristics of the data inform the tasks to be performed and have a direct impact not only on the predictive model, but also on the final optimization model. For example, the amount of data available might influence the choice of the ML method used to learn the constraint, which may then have an influence on the approach used to solve the final optimization model. If the right data are available, then they can be used directly (with or without preprocessing) to learn the constraints \eqref{eqn:step1Epred}. If the data are not yet available, decisions have to be made on how to collect the data. Data collection or generation strategies should follow established principles such as Design of Experiments (DoE) techniques (which are applicable for physical experiments) or Design of Computer Experiments (DoCE) techniques (applicable for computational experiments). In the WFP application of Section \ref{sec:WFPExample} for example, collecting data on the palatability of rations during an epidemic or in a war zone might not be possible. In this case, the generation or simulation of data following DoCE principles can be done. A recent review of DoCE is given in \cite{GARUD201771}.

Data collection or generation can take place either in a single run or in multiple runs. Performing data collection multiple times can make the learning and optimization process more accurate. For example, if the simulation of a process from which data is obtained takes several hours, then the choice of input parameters for the simulator might need to be carefully considered. Adaptive DoCE techniques such as those in \cite{CROMBECQ2011683} and \cite{xu2014robust} may be used in such cases. In Derivative-Free Optimization (DFO) only a few or even one new data point is collected in each iteration to minimize the number of data points needed to obtain an optimal solution. We refer to \cite{Conn2009IntroductionTD} for a detailed treatment on DFO. 
In addition to the availability or sourcing of the data, the size of the dataset also informs the choice of the learning approach. It is known that deep neural networks are able to exploit large datasets better than traditional algorithms such as logistic regression or support vector machines \citep{NgDeepLearn2020}. In contrast, methods such as kriging \citep{matheron1963principles} can be used with smaller datasets. Regardless of the size of the datasets, most data (especially those not generated with the use of simulators) need to be pre-processed before they can be used in the next step (learning). Examples of various pre-processing methods are given in \cite{mckinney2017python}. Whatever data collection, generation, and preprocessing approaches are used, the end result of this second step is a dataset that is ready to be used for training, testing, and validating the predictive algorithms. A review of the various data generation and preprocessing approaches used in the context of constraint learning is given in Section \ref{sec:dataHandlingReview}.

\subsection{Step 3: Selection and Training of Predictive Model(s)}
\label{sec:overviewStep3}
Once the data gathering and preprocessing activities have been completed, it is necessary to select, train, validate, and test the predictive model that will subsequently be used to represent one or more constraints in the final optimization model. Selection and training are two activities that go hand in hand. It is not possible to select the best predictive model without first training and testing it on the available dataset. Correspondingly, the poor performance of a model after training might require a different predictive model to be selected. The selection of the predictive model is based on six main factors:
\begin{enumerate}
    \item[i]Classification or regression: The choice of the predictive model will depend on whether the constraint to be learned results in a discrete set of values (classification), or a continuous spectrum of values (regression). The use of a classification model results in a feasible set being learned, while the use of a regression model results in a function being learned. 
    
    \item[ii] Computational complexity: The choice of the predictive model directly affects the computational complexity of the final optimization model. In contrast to a linear regression model, deep learning models require several mathematical or logical statements to represent them (see Section \ref{sec:modelResolveReview}) and undoubtedly increase the complexity of the final optimization problem. A decision might also be made to select a predictive model so that the complexity of the final optimization problem does not increase. In the WFP example, the underlying problem is linear so a linear predictive model may be chosen to learn \eqref{eqn:palatability}.
    
    \item[iii] Model performance: In addition to computational complexity, the predictive model has to perform adequately with respect to some performance measure (e.g. accuracy or mean squared error). A simple but inaccurate model can have worse consequences than a complex but accurate model. The performance should be evaluated on both a test set and in the final optimization model. This means that the performance evaluation has to be performed both in step 3 and step 5. 
    
    \item[iv] Confidence intervals: If the predictive models used provide confidence intervals, it might be possible to use this information in approaches such as stochastic optimization or robust optimization.
    
    \item[v] Data quality: Despite the attention focused on the dataset in Step 2, it may be necessary to choose a predictive algorithm that is robust to noise or missing data. This is especially the case if the data to be used has been provided by a third party.
    
    \item[vi] Interpretability: The selection of a predictive model might also depend on whether or not the practitioner requires the model to be interpretable. In the radiotherapy optimization problem of Section \ref{sec:RadOptExample}, clinicians may prefer highly explainable predictive models to be used to learn the $\text{TCP}(\bm{d}, \bm{\gamma})$ and $\text{NTCP}_q(\bm{d}, \bm{\gamma})$ functions due to the seriousness of the application. There are however pros and cons to consider. Decision trees, for example, are known to be highly explainable but may require the linearization of several disjunctions to encode them in the optimization problem. Refer to Section \ref{sec:modelResolveReview} for an example.
\end{enumerate}

Table \ref{tab:overview} shows the pros and cons of the predictive models generally used in constraint learning. It can be seen that while linear models are very good in terms of explainability and computational complexity, their performance is lacking. Conversely, while neural networks perform quite well, they require large amounts of labeled data to train \citep{NgDeepLearn2020} and are not considered explainable \citep{molnar2020interpretable}. A review of the various predictive models used for constraint learning in optimization is given in Section \ref{sec:modelSelectImplReview}.

\begin{table}[h]
\caption{Pros and cons of predictive models commonly used in constraint learning}
\label{tab:overview}
\resizebox{\textwidth}{!}{%
\begin{tabular}{@{}lcccc@{}}
\toprule
& Explainability & Complexity & Data Required & Performance \\ \midrule
Linear models$^*$  & ++ & ++ & ++ & \texttt{-{}-} \\
\rowcolor{gray!10} Decision Trees &  + & + & + & + \\
Tree ensembles  & \texttt{-} & \texttt{-} & \texttt{-} & + \\
\rowcolor{gray!10} Neural Networks &  \texttt{-{}-} & \texttt{-{}-} & \texttt{-{}-} & ++ \\ \bottomrule
\multicolumn{4}{l}{\footnotesize ++: very good; +: good; \texttt{-}: bad; \texttt{-{}-}: very bad}\\
\multicolumn{4}{l}{\footnotesize $^*$  Linear regression, linear support vector machine}
\end{tabular}}
\end{table}

The implementation process is a cycle of training the predictive model, evaluating the model's performance, and improving the model in terms of the data and the model parameters. If a feed-forward neural network is used, for example, parameters such as the number of layers, the number of nodes per layer, the activation functions, whether or not to use dropout regularization, etc., need to be determined. Out-of-sample testing should be done to ensure the adequate performance of the model, with changes made to improve performance if necessary. In the context of constraint learning, under the same computational complexity, the best predictive model is the one with higher performance in the final optimization model. This is related to the concept of predictive versus prescriptive models and is explained further in \cite{bertsimas2020predictive}. The goal should be to generate prediction models that aim to minimize decision error, and not just prediction error \citep{elmachtoub2022smart}.

\subsection{Step 4: Resolution of the Optimization Model}
\label{sec:overviewStep4}
Once the predictive model has been used to learn the difficult-to-model constraint, it will need to be embedded into the rest of the optimization problem. Recall that the predictive model $\hat{\bm{h}}(\mathord{\cdot})$ in \eqref{eqn:step1Epred} may not be easy to add as a constraint and may need to be encoded in such a way that it can be handled by commercial or conventional solvers. This however depends on the type of model used, as a linear or quadratic function can be embedded more easily than a neural network for example. The integration of the learned constraints is often not trivial and may result in a significant reformulation of the model. If possible, the predictive model may be incorporated either by providing gradients directly to the solver, or by reformulating it such that it becomes linear, convex, conic, etc., and can more readily be incorporated into the optimization problem. If desired, the optimization models may be modified to take into account any uncertainty measures provided by the predictive models. A successful embedding can also include providing additional information (such as bounds) to the solver of choice in order to facilitate a more efficient solution process. A successful embedding should make it possible to take full advantage of the optimization solver's capabilities. A deeper look at the various methods used in the literature to encode predictive models is given in Section \ref{sec:modelResolveReview}.

\subsection{Step 5: Verification and Improvement of the Optimization Model}
\label{sec:overviewStep5}
In this final step, the performance of the constraint learning process is evaluated with respect to three things: the final optimization model, the predictive model, and the data. For the optimization model, the focus is on both the goodness of the solution (i.e. the objective function value), as well as the time taken to find this solution. For the performance of the predictive model, the focus is on ensuring that the learned constraints accurately represent the difficult-to-model constraints. The data used to learn constraints must also be examined. If, for example, the data points used are far from the boundary, this might have a negative effect on the constraint learned and the optimal solution \citep{thams2017data}. If it is possible to generate data, then using different data to improve the predictive model can in turn help to improve the optimal solution. One way of doing this is via a process called active learning \citep{settlesTr09}. Here, the dataset is continually modified by adding new solutions (output of the optimization model) to it, which is then used to re-train the predictive model. A review of the approaches used to verify and improve models with learned constraints is provided in Section \ref{sec:modelVerifReview}.\\ 

 Table \ref{tab:juxtaposition} shows the emphasis placed on the different steps in the literature. While all steps are generally present in the literature, some steps are often emphasized more than others. It can be seen that the most emphasis is placed on Step 4, i.e., learning and reformulating constraints so that the final optimization problem can be easily solved.  While the above framework steps have been given in sequential order, it should be noted that some of these steps may be repeated. For example, if it is seen in step 5 that the learned constraint is a poor representation of reality, we may need to go back to step 1. Similarly, the incorporation of new data will mean that the predictive model needs to be retrained and re-embedded, and so on. Internal loops are also possible --- it may be necessary to iterate between steps 2 and 3 until an acceptable predictive model is developed, or between steps 4 and 5 until satisfactory results are achieved. 

\rowcolors{3}{gray!10}{white}
\begin{table}
\centering
\caption{Emphasis placed on the different steps of the framework}
\resizebox{\textwidth}{!}{\begin{tabular}{ @{} l *{6}{c} @{} } 
\toprule
  & Step 1 & Step 2  & Step 3 & Step 4 & Step 5 \\
\midrule
\cite{bergman2022janos} & \CIRCLE  & \CIRCLE & \CIRCLE & \CIRCLE & \CIRCLE \\
\cite{biggs2018optimizing} & \CIRCLE & \CIRCLE & \CIRCLE & \CIRCLE & \CIRCLE \\
\cite{chen2020input} & \LEFTcircle & \LEFTcircle & \CIRCLE & \CIRCLE & \LEFTcircle  \\
\cite{chi2007modeling} & \Circle & \CIRCLE & \CIRCLE & \CIRCLE & \CIRCLE \\
\cite{cozad2014learning} & \LEFTcircle & \CIRCLE & \CIRCLE & \CIRCLE & \CIRCLE  \\
\cite{cremer2018data} & \LEFTcircle & \LEFTcircle & \CIRCLE & \CIRCLE & \CIRCLE \\
\cite{de2003integrating} & \CIRCLE & \CIRCLE & \CIRCLE & \LEFTcircle &  \CIRCLE \\
\cite{fahmi2012process} & \CIRCLE & \CIRCLE & \CIRCLE & \CIRCLE & \LEFTcircle \\
\cite{garg2018kernel} & \CIRCLE & \LEFTcircle & \CIRCLE & \CIRCLE & \CIRCLE \\
\cite{grimstad2019relu} & \CIRCLE & \CIRCLE & \CIRCLE & \CIRCLE &  \CIRCLE  \\
\cite{gutierrez2010neural} & \CIRCLE & \LEFTcircle & \CIRCLE & \CIRCLE & \CIRCLE  \\
\cite{halilbavsic2018data} & \LEFTcircle & \CIRCLE & \CIRCLE & \CIRCLE & \CIRCLE  \\
\cite{jalali2019designing} & \CIRCLE & \CIRCLE & \CIRCLE & \CIRCLE & \CIRCLE \\
\cite{karmelita2020cma} & \CIRCLE & \CIRCLE & n/a & \CIRCLE & \CIRCLE  \\
\cite{kudla2018one} & \CIRCLE &  \CIRCLE & \CIRCLE & \CIRCLE & \CIRCLE  \\
\cite{kumar2019automating} & \CIRCLE & \CIRCLE & n/a & \CIRCLE & \CIRCLE \\
\cite{kumar2019acquiring} & \CIRCLE & \CIRCLE & n/a & \CIRCLE & \CIRCLE \\
\cite{kumar2021learning} & \CIRCLE & \CIRCLE & n/a & \CIRCLE & \CIRCLE  \\
\cite{lombardi2017empirical} & \CIRCLE & \CIRCLE & \CIRCLE & \CIRCLE & \CIRCLE  \\
\cite{maragno2021mixedinteger} & \CIRCLE & \CIRCLE  & \CIRCLE & \CIRCLE & \CIRCLE  \\
\cite{mivsic2020optimization} & \CIRCLE & \LEFTcircle & \CIRCLE & \CIRCLE & \CIRCLE \\
\cite{venzke2020neural} & \CIRCLE & \CIRCLE & \CIRCLE & \CIRCLE & \CIRCLE \\
\cite{paulus2021comboptnet} & \CIRCLE & \CIRCLE & \CIRCLE & \CIRCLE & \CIRCLE \\
\cite{pawlak2017automatic} & \CIRCLE & \LEFTcircle & n/a & \CIRCLE & \CIRCLE  \\
\cite{pawlak2017synthesis} & \CIRCLE & \CIRCLE & n/a & \CIRCLE & \CIRCLE \\
\cite{pawlak2018synthesis} & \CIRCLE & \CIRCLE & n/a & \CIRCLE & \CIRCLE \\
\cite{pawlak2019synthesis} & \CIRCLE & \CIRCLE & n/a & \CIRCLE & \CIRCLE \\
\cite{pawlak2021ellipsoidal} & \CIRCLE & \CIRCLE & \CIRCLE & \CIRCLE & \CIRCLE \\
\cite{pawlak2021grammatical} & \CIRCLE & \CIRCLE & n/a & \CIRCLE & \CIRCLE  \\
\cite{prat2020learning} & \LEFTcircle & \CIRCLE & \CIRCLE & \CIRCLE & \LEFTcircle  \\
\cite{say2017nonlinear} & \LEFTcircle & \CIRCLE & \CIRCLE & \CIRCLE & \LEFTcircle   \\
\cite{schede2019learning} & \CIRCLE & \Circle & \CIRCLE & \CIRCLE & \CIRCLE  \\
\cite{schweidtmann2019deterministic} & \LEFTcircle & \LEFTcircle & \CIRCLE & \CIRCLE & \LEFTcircle \\
\cite{spyros2020decision} & \CIRCLE & \CIRCLE & \CIRCLE & \CIRCLE  & \CIRCLE  \\
\cite{sroka2018one} & \CIRCLE & \CIRCLE & n/a & \CIRCLE & \CIRCLE \\
\cite{thams2017data} & \LEFTcircle & \CIRCLE & \CIRCLE & \CIRCLE & \CIRCLE \\
\cite{verwer2017auction} & \CIRCLE & \LEFTcircle & \LEFTcircle & \CIRCLE & \CIRCLE \\
\cite{xavier2021learning} & \LEFTcircle & \CIRCLE & \CIRCLE & \CIRCLE  & \CIRCLE \\
\cite{yang2021optimization} & \CIRCLE & \CIRCLE & \CIRCLE & \CIRCLE & \CIRCLE \\
\bottomrule
\rowcolor{white}\multicolumn{4}{l}{\footnotesize \CIRCLE: most emphasis; \LEFTcircle: some emphasis; \Circle: less emphasis; n/a: not applicable}\\
\end{tabular}}
\label{tab:juxtaposition}
\end{table}

      \section{Review}
\label{sec:review}
\subsection{Setup of the Conceptual Model}
\label{sec:modelSetupReview}
The papers by \cite{lombardi2017empirical} and \cite{maragno2021mixedinteger} provide very detailed approaches to formalizing the process of OCL. There is also a clear structure provided in \cite{de2003integrating} for learning unknown constraints. The methodology in \cite{lombardi2017empirical}, called Empirical Model Learning (EML), is described as a way of obtaining components of a prescriptive model from machine learning. Similar to the steps given in Section \ref{sec:overviewStep1}, they define a conceptual model including known constraints, constraints to be learned, as well as the definition of constraints used for feature extraction. \cite{maragno2021mixedinteger} also provide a general framework for OCL that can be applied every time data is available. A key focus of their approach is to learn constraints and embed them in such a way that the computational complexity of the final model is not increased. \cite{padmanabhan2003use} and \cite{deng2018coalescing} also conceptually discuss the use of data mining to define objectives and constraints.

While other papers which use a predictive element to learn constraints do not have as detailed a setup, they incorporate some of the elements of Step 1 in their approach. Some authors (e.g. \cite{verwer2017auction}) explicitly define constraints to perform feature extraction, while most of the papers reviewed simply identify which constraints are difficult to model manually and replace them with predictive models. This will be discussed later in Sections \ref{sec:modelSelectImplReview} and \ref{sec:modelResolveReview}.

The approach taken to formalize the constraint learning process in \cite{pawlak2017automatic} is different in that their framework is a MILP. Similarly, other authors also leverage MILP \citep{schede2019learning} or Integer Programming (IP) \citep{cozad2014learning} frameworks for constraint learning. \cite{sroka2018one} use local search, and \cite{kumar2021learning} use an approach based on stochastic local search. A number of authors also use a genetic programming framework to obtain constraints from data \citep{pawlak2017synthesis,pawlak2018synthesis}. \cite{pawlak2019synthesis} and \cite{pawlak2021grammatical} use evolutionary strategies. In these different setups, the authors give the relative simplicity of their approaches as their key justification  --- no predictive models have to be learned, no constraints for feature extraction have to be included, and no complex embedding or reformulating procedures need to be followed. The disadvantage of these approaches however is that they do not make use of the learning power of predictive models which have been shown to  be able to learn very complex decision boundaries well.


\subsection{Data Gathering and Preprocessing}
\label{sec:dataHandlingReview}
Of the papers considered in this survey, \cite{thams2017data} has a detailed approach with regards to the data generation and preprocessing phase detailed in Section \ref{sec:overviewStep2}. Their procedure, also used by \cite{halilbavsic2018data} and \cite{venzke2020neural}, is designed to either make use of available data from sensors or generate them using a simulator. Possible operating points of the system under consideration are fed into a simulator, and the output of the simulator is analyzed and classified as either \textit{stable} or \textit{unstable}. These operating points along with their classifications are then stored in a database. The data is only generated once, and they attempt to keep the generated database as small as possible for computational reasons. Although the generated data is small, they ensure that enough data is collected so that their predictive models can be trained. They do this by focusing on generating points close to the boundary of their two-class problem while neglecting operating points far away from the boundary. In this way, they balance exploration and exploitation when generating data. In addition to this, class imbalance in the dataset is addressed using several strategies, and feature selection is done so that only easily accessible variables are considered.

With regards to following DoCE principles, the authors of \cite{lombardi2017empirical} use factorial designs to generate multiple training sets from a simulator. Data generation here, although time-consuming, only needs to be done once. The data is also normalized so that the input features fall within a certain range. The approach developed in \cite{cozad2014learning} also takes DoCE principles into account by using either Latin Hypercube Sampling (LHS) or two-level factorial design \citep{mukerjee2006modern}. They also use adaptive sampling to iteratively refine the models generated and provide discussions on whether to use exploration-based or exploitation-based sampling. \cite{de2003integrating} use small data collected using an on-field survey (an expensive process) to find a probability distribution that is later used in a simulator that generates enough data to train the predictive model. In \cite{fahmi2012process}, the data is obtained from multiple simulators. Some transformation is done to the data to ensure that the distribution of the dependent variable is as close as possible to a uniform distribution. The authors here note that a large amount of data was needed to train their neural networks for higher accuracy. Data may also be collated from multiple available sources as in the case of \cite{bertsimas2016analytics}.

In terms of the sizes of the datasets surveyed, it can be seen that not many of the problems considered have large datasets. Data containing $500$ or fewer examples are common. In \cite{pawlak2017automatic}, the total amount of data generated using uniform sampling varied from $10$ to $500$ examples. Similarly, \cite{schweidtmann2019deterministic} use data with $500$ or fewer examples. Data of a similar size is also seen in \cite{kudla2018one}, \cite{yang2021optimization} and several others. For one of the test problems in \cite{schweidtmann2019deterministic}, only $46$ examples are used. On the larger side, \cite{chen2020input} use data containing $10,000$ instances, \cite{venzke2020neural} use $36,144$ data points generated using LHS, while the data used in \cite{say2017nonlinear} contains $100,000$ data points. For certain constraint learning approaches, the availability of a large amount of data is a must. The local search approach used in \cite{sroka2018one} is such an example, where the test set contains $500,000$ examples generated via uniform sampling. 

Some approaches for constraint learning necessitate the significant preprocessing of the data. In order to learn ellipsoidal constraints in \cite{pawlak2021ellipsoidal}, the uniformly generated data first needs to be clustered and standardized before Principal Component Analysis (PCA) is applied to the data. This clustering and PCA procedure shapes the data so that it can be represented using ellipsoidal constraints. In other cases, although preprocessing is done, it is not crucial to the constraint learning approach (e.g. in \cite{biggs2018optimizing} and \cite{pawlak2019synthesis}). 

In some cases, not much emphasis is placed on the data generation and preprocessing phase. This is sometimes because the data is fully available, is neither big nor dirty, and has no need for preprocessing. For example, the situation of having high-dimensional data, where the number of examples is much less than the number of features, is only explicitly mentioned in \cite{mivsic2020optimization}. In a few cases, however, the data step is an afterthought. Even when data is generated, this is cursorily mentioned without giving details or following exhaustive DoCE principles. We believe that our framework will help to provoke more thought on this phase of the constraint learning process.


\subsection{Selection and Training of ML Models}
\label{sec:modelSelectImplReview}
\rowcolors{3}{gray!10}{white}
\begin{sidewaystable}
\centering
\caption{Methods used for constraint learning}
\resizebox{\textwidth}{!}{\begin{tabular}{ @{} l *{8}{c} @{} } 
\toprule
  & Neural & Decision  & Random & Other & Support Vector & Clustering & (M)ILP & Other\\
  & Networks & Trees  & Forest & Ensemble & Machines & &  & \\
\midrule
\cite{bergman2022janos} & x & &  &  & & &  & x \\
\cite{biggs2018optimizing} &  & x & x &  & & &  &  \\
\cite{chen2020input} & x &  &  &  &  &  &  & \\
\cite{chi2007modeling} &  &  &  &  & x &  &  &  \\
\cite{cozad2014learning} &  &   &  &  &  &  & x & x \\
\cite{cremer2018data} &  & x &  & x & & &  &  \\
\cite{de2003integrating} & x &  &  &  &  &  &  &  \\
\cite{fahmi2012process} & x &  &  & & &  &  & x \\
\cite{garg2018kernel} &  &  &  &  & x &  &  &  \\
\cite{grimstad2019relu} & x &  &  &  &  &  &  &  \\
\cite{gutierrez2010neural} & x &  &  & & &  &  &  \\
\cite{halilbavsic2018data} &  & x &  &  & & &  &  \\
\cite{jalali2019designing} &  &  &  &  & x &  &  &  \\
\cite{karmelita2020cma} &  &  &  &  &  & x &  & x \\
\cite{kudla2018one} &  & x &  &  &  &  & & \\
\cite{kumar2019automating} &  &  &  &  &  &  &  & x \\
\cite{kumar2019acquiring} &  &  &  &  &  &  & x & x \\
\cite{kumar2021learning} &  &  &  &  &  &  &  & x \\
\cite{lombardi2017empirical} & x & x &  &  &  &  &  &  \\
\cite{maragno2021mixedinteger} & x & x  & x & x & x & x &  & x \\
\cite{mivsic2020optimization} &  & x & x &  &  &  & &  \\
\cite{venzke2020neural} & x &  &  &  &  &  &  &  \\
\cite{paulus2021comboptnet} & x &  &  &  &  &  &  &  \\
\cite{pawlak2017automatic} &  &  &  &  &  &  & x &  \\
\cite{pawlak2017synthesis} &  &  &  &  &  &  &  & x \\
\cite{pawlak2018synthesis} &  &  &  &  &  &  &  & x \\
\cite{pawlak2019synthesis} &  &  &  &  &  &  &  & x \\
\cite{pawlak2021ellipsoidal} &  &  &  &  &  & x &  & x \\
\cite{pawlak2021grammatical} &  &  &  &  &  &  & x & x \\
\cite{prat2020learning} &  & x &  &  &  &  &  &  \\
\cite{say2017nonlinear} & x &  &  &  &   &  &  &  \\
\cite{schede2019learning} &  & x &  &  &  &  & x &  \\
\cite{schweidtmann2019deterministic} & x &  &  &  &  &  &  &  \\
\cite{spyros2020decision} & x & x &  &  &  &  &  &  \\
\cite{sroka2018one} &  &  &  &  &  & x &  & x \\
\cite{thams2017data} &  & x &  &  &  &  &  &  \\
\cite{verwer2017auction} &  & x &  &  &  &  &  & x \\
\cite{xavier2021learning} &  &  &  &  & x &  &  &  \\
\cite{yang2021optimization} & x &  &  &  &  &  &  &  \\
\bottomrule
\end{tabular}}
\label{tab:methodsUsed}
\end{sidewaystable}

We first note that papers that use regression functions as the main constraint learning methods have been excluded from Table \ref{tab:methodsUsed}, as they are too numerous to be included in this paper. Regression with first- and second-order polynomials is used a lot in engineering, especially in combination with DoE and DoCE. We refer to \cite{kleijnen2015} for a detailed treatment. Kriging models are also used a lot in engineering, especially when the objective function values can be obtained by (expensive) computer simulations. The big advantage of kriging is that it has very high predictive power even for a relatively small number of data points. The disadvantages of kriging are (i) it costs much computing time to obtain the kriging model (ii) the data points have to be space-filling (iii) embedding kriging functions in the optimization model makes it non-convex and hard to solve. See \cite{forrester2008} for a detailed treatment on kriging.

It can be seen from Table \ref{tab:methodsUsed} that Artificial Neural Networks (ANN) and Decision Trees (DT) are the most popularly used predictive models in constraint learning (with the exception of regression models). This could be because these methods have been shown to have high representative power, can be used for both regression and classification, and generally perform well. Reviews on ANNs and DTs are given in \cite{cao2018review} and \cite{carrizosa2021mathematical} respectively. Training, testing, and evaluation are generally done as prescribed in the ML literature, and the methods are generally evaluated using metrics such as the Mean Square Error (MSE), accuracy, and so on. For example, to learn objective functions in \cite{biggs2018optimizing}, the data is split in a 70:30 ratio between training and test sets, and prediction accuracy used as the evaluation metric. See Chapter 2 of \cite{james2013introduction} for more information on training and testing ML models.

In terms of the size of the ANNs, with the exception of \cite{yang2021optimization} who use a network with six hidden layers, most networks used in constraint learning are generally small. One reason for this could be to reduce the burden of embedding large networks as constraints --- it can be seen from \eqref{eqn:reluA} to \eqref{eqn:reluE} that the number of binary variables required is proportional to the size of the network. Also, as the input of a current layer is the output of a previous one, using multiple layers results in nested activation functions in the final model. This could lead to issues regarding convexity and/or differentiability, therefore smaller networks are used. Another reason for small neural network sizes is that using larger networks provides no benefit in some cases. \cite{lombardi2017empirical} for example use a network with one hidden layer (with two neurons) as increasing the network size did not improve the solution quality of the optimization problem. To view the effect of network size on computation times, \cite{schweidtmann2019deterministic} vary both the number of neurons and the number of layers. Results showed that the solution time increases approximately exponentially with the number of layers used to learn the constraints. Deep neural networks required more processing time than shallow neural networks for the same total number of neurons.

An alternative to using small networks could be to use sparsification. \cite{venzke2020neural} use a network with three hidden layers and fifty neurons in each layer to learn constraints, however they reduce the size of the network by enforcing $70\%$ of the weights to be zero. This sparsification procedure ensures that the network was not too large and also avoided over-fitting during training. \cite{say2017nonlinear} use a similar technique to remove connections to and from neurons with very small weights in order to strengthen the MILP formulation. Another alternative could be to ensure that the constraint learning procedure results in a convex optimization problem. \cite{chen2020input} and \cite{yang2021optimization} use Input Convex Neural Networks (ICNN) \citep{amos2017input} as their predictive models. An ICNN requires that all weights in the network are non-negative and that all activation functions are convex and non-decreasing. Although these networks require more training iterations and have lower representation power than conventional networks, their use guarantees that the final problem is a convex optimization problem. This allows the authors to balance the trade-off between model accuracy and computational tractability.

The activation functions used in the hidden layers in the constraint learning literature are usually either the Rectified Linear Unit (ReLU) (e.g. in \cite{say2017nonlinear}), the sigmoid function (e.g. in \cite{chen2020input}),  (some form of) the hyperbolic tangent function (e.g. in \cite{fahmi2012process}), or the softplus activation function (e.g. in \cite{chen2020input}). The output layer activation function is usually an identity linear function. An overview of the most common types of activation functions is given in \cite{sharma2017}. 

In addition to ANNs, DTs are also popularly used in constraint learning. DTs are able to capture non-convex and discontinuous feasibility spaces \citep{spyros2020decision}, and provide these rules in an easily understandable format \citep{prat2020learning}. For this reason, \cite{thams2017data} and \cite{halilbavsic2018data} specifically choose decision trees as their method of choice to learn constraints in an optimal power flow problem. Their rationale is that as electricity is a crucial utility, using easily interpretable rules in their analysis can lower the barrier to power plant operators adopting their methods. This interpretability requirement could also favor the use of decision trees in the radiotherapy example of Section \ref{sec:RadOptExample}.

As random forests are a natural extension of using decision trees, they have also been used in constraint learning. \cite{biggs2018optimizing} learn objective functions using random forest while \cite{mivsic2020optimization} optimize tree ensembles. AdaBoost \citep{freund1997decision} (with DTs as base learners) is also used in \cite{cremer2018data} to learn a probabilistic description of a constraint. 

Support Vector Machines (SVM) can be expressed very clearly as optimization problems, and as a result, there is a significant amount of optimization literature on SVMs. Most of the literature however deals with subjects such as optimal feature selection (e.g. \cite{JIMENEZCORDERO202124}), or mathematical formulation and solution of the SVM problem (e.g. \cite{BALDOMERONARANJO202084}), or improving predictions (e.g. \cite{YAO2017679}). There are however a few cases where SVM is used in the constraint learning process. \cite{jalali2019designing} use SVM to learn power inverter control rules. \cite{chi2007modeling} use SVM for regression with a second order polynomial kernel function. A feature selection process is further used to remove insignificant terms and improve model accuracy. \cite{xavier2021learning} also use SVM with linear kernels to learn hyper-planes within which the solution of the optimization problem is very likely to exist. 

Clustering is also used in constraint learning, although it is often used in addition to other methods. \cite{pawlak2021ellipsoidal} combine clustering with PCA and ellipsoidal constraints to get an MIQCP, while \cite{sroka2018one} combine clustering with local search to search the clusters for LP constraints that represent each cluster.

Combinations of predictive models are often used. \cite{verwer2017auction} combine regression trees and regression models  for learning revenue constraints for an auction optimization problem. In addition to using ANNs to learn most of the constraints required, \cite{fahmi2012process} also use non-linear regression models for specific constraints. \cite{paulus2021comboptnet} use both integer programming and neural networks to learn both the cost terms and the constraints for integer programs. Although \cite{schede2019learning} use an LP formulation to learn polytopes that cover feasible examples, they also include a decision tree heuristic to model the importance of examples to be considered. Additionally, \cite{maragno2021mixedinteger} are able to use different predictive models to learn different constraints of the same optimization problem in order to achieve better performance. 

There are also approaches that do not use predictive models to learn constraints. \cite{pawlak2017synthesis,pawlak2018synthesis} and \cite{pawlak2019synthesis} use genetic programming as their approach of choice, \cite{pawlak2017automatic} use a mixed-integer formulation and \cite{cozad2014learning} combine simple basis functions using a MILP formulation to learn constraints. 

Although RF is an ensemble method, the column ``Other Ensemble" in Table  \ref{tab:methodsUsed} is separated from the ``Random Forest" column to highlight the fact that there is an opportunity to use other ensemble methods where different individual base learners can be used and their predictors combined. There is also the opportunity to try a wider variety of predictive models such as symbolic regression or Bayesian methods, especially if they are more easily represented using mathematical functions.  Studies could be done to see if certain families of models perform better for certain types of problems. Finally, methods that provide some measure of uncertainty (such as confidence intervals or conditional probabilities) might also be chosen in order to be able to incorporate this uncertainty into the final mathematical model.

\subsection{Resolution of the Optimization Model}
\label{sec:modelResolveReview}
With regression functions, the process of embedding them as constraints is relatively straightforward as the functions are simply a weighted sum of the decision variables \citep{bertsimas2016analytics,verwer2017auction}. With other predictive models, however, the embedding process is more involved. With ANNs for example, their embedding is also dependent on the activation functions used. The output $\hat{y}$ of a fully connected neural network with $h$ hidden layers can be written as:
\begin{subequations}
\begin{align} 
\bm{\mathrm{H}}_1 = \sigma_1 \left( \bm{\mathrm{W}}_1 \bm{\mathrm{X}} \right) \label{eqn:fcnnA} \\
\bm{\mathrm{H}}_i = \sigma_i \left( \bm{\mathrm{W}}_i \bm{\mathrm{H}}_{i-1} \right) \label{eqn:fcnnB} \\
\hat{y} = \sigma_{h+1} \left( \bm{\mathrm{W}_{h+1}} \bm{\mathrm{H}}_{h} \right), \label{eqn:fcnnC}
\end{align}
\end{subequations}
where $\bm{\mathrm{X}}$ is the training data, $\bm{\mathrm{W}}_i$ are the weights associates with layer $i$, and $ \sigma_i$ is the activation function used in in layer $i$. In order to incorporate a trained neural network with ReLU activation functions (i.e. $\sigma(x) = max(0,x)$) as a constraint, the $max()$ function can be replaced by introducing binary variables $b_j$ for each neuron in the network: 
\begin{subnumcases}{y_j = max \left(0, \hat{y}_j \right) \implies}
  y_j \leq \hat{y}_j - \hat{y}_j^{min} \left( 1-b_j \right) \label{eqn:reluA} \\
  y_j \geq \hat{y}_j \label{eqn:reluB} \\
  y_j \leq \hat{y}_j^{max} b_j \label{eqn:reluC} \\
  y_j \geq 0 \label{eqn:reluD} \\
  b_j \in \{0,1\}^{N_j}. \label{eqn:reluE}
\end{subnumcases}

If the input to a neuron $\hat{y}_j > 0$, then $b_j = 1$ and \eqref{eqn:reluA} and \eqref{eqn:reluB} force the neuron output $y_j$ to the input $\hat{y}_j$. On the other hand, if the input to the neuron $\hat{y}_j\leq 0$ then $b_j = 0$, and \eqref{eqn:reluC} and \eqref{eqn:reluD} force the output $y_j$ to $0$. The resulting MILP can then be solved with off-the-shelf solvers such as Gurobi \citep{gurobi} or CPLEX \citep{cplex}. This is done by authors such as \cite{venzke2020neural} and \cite{bergman2022janos}. 

The disadvantages of this approach are twofold. Firstly, large networks require a correspondingly large number of binary variables. Secondly, the bounds $\hat{y}_j^{min}$ and $\hat{y}_j^{max}$ have to be chosen to be as tight as possible for the MILP solver. \cite{anderson2020strong} present a MIP formulation that produces strong relaxations for ReLUs. They start with the big-M formulation and then add strengthening inequalities as needed. \cite{say2017nonlinear} also encode ReLU into MILP using big-M constraints, but include reformulations to strengthen the bounds. They also use sparsification to remove neurons with very small weights, as constraints with very small coefficients can be difficult to solve for commercial solvers \citep{d2016towards}.

The embedding of trained ANNs as constraints is not always as complicated as mentioned above. Equations \eqref{eqn:fcnnA} to \eqref{eqn:fcnnC} can be used directly as constraints, as long as the solver can handle the activation function $\sigma$. This approach is used in \cite{fahmi2012process} with the hyperbolic tangent activation function, as well as in \cite{gutierrez2010neural} where the activation function is differentiable so that the constraint can be easily handled by conventional solvers. It is nevertheless noted in \cite{lombardi2017empirical} that several commercial solvers rely on convexity for providing globally optimal results and that this approach may result in locally optimal solutions depending on the $\sigma$ used. In \cite{chen2020input}, the use of ICNN guarantees that the optimization problem is convex and will converge to a globally optimal solution. They do however speed up this process by using a smoothened version of the ReLU, called the Softplus activation function. A large value of the parameter $t$ in the Softplus function allows it to be smooth enough, while still being convex. The final problem is solved using dual decomposition \citep{yu2006dual,chiang2007layering}. \cite{yang2021optimization} also use ICNNs to get convex optimization problems which are then solved using sequential quadratic programming.

Noticing the use of ANNs in various optimization applications, \cite{schweidtmann2019deterministic} provide an efficient method for deterministic global optimization problems which have ANNs embedded in them. As the only non-linearities of ANNs are their activation functions (the $\tanh$ function in their case), the tightness of their relaxations is influenced by the relaxations of the $\tanh$ functions. The key idea of their paper is to therefore hide the equations that describe the ANN and its variables from the Branch-and-Bound (B\&B) solver using McCormick relaxations of the activation functions. These relaxations are once continuously differentiable, and the lower bounds are calculated by automatic propagation of McCormick relaxations through the network equations. This ``Reduced-Space (RS)" formulation results in significantly fewer variables and equalities than the original formulation (by two orders of magnitude in some cases), with significantly shorter computation times in most cases. They also see that shallow ANNs show much tighter relaxations than deep ANNs. \cite{grimstad2019relu} also provide bound tightening procedures for ReLU networks to reduce solution times.

The embedding of kriging functions in the final model is done in a direct way: substituting the kriging function leads to an explicitly formulated, but highly non-convex constraint. First and second-order derivatives can  be calculated, and therefore standard nonlinear solvers can be used. However, due to the non-convexity, solvers might end up in local optima. See also \cite{dbb011afe7154a63a10e75e379fa5a1f}, and \cite{STINSTRA2008816} for examples. 

With DTs, the paths from the root to leaf nodes can be represented using disjunctions of conjunctions. These rules can either be linearized or embedded as constraints using a framework like Generalized Disjunctive Programming (GDP) \citep{Grossmann2009,GAMS2021}. Although \cite{lombardi2017empirical} do not embed DTs into MINLPs due to the potential for poor bounds being obtained as a result of the extensive linearization of disjunctions required, other authors have. In \cite{kudla2018one}, branches of the tree from root to leaf nodes are encoded as linear constraints using a big-M formulation. Paths that end up in feasible decisions are denoted with the (+) sign (see Figure \ref{fig:decTreeEmbed}). These paths are then converted into linear constraints using auxiliary binary variables to show which path in the tree is followed. For example, at the top of the tree, if a binary variable $b_1=1$, then $c_1$ is active and \eqref{eqn:c1} holds. Otherwise if $b_1=0$, then $c_2$ is active and \eqref{eqn:c2} holds. Mathematically,
\begin{align}
    c_1: & \, x_1 \leq a_1 \rightarrow \, x_1 \leq a_1 + M \left( 1 - b_1 \right) \label{eqn:c1}\\
    c_2: & \, x_1 > a_1 \rightarrow \, x_1 > a_1 - M b_1. \label{eqn:c2}
\end{align}

\begin{figure}[hbtp]
\centering
\includegraphics[scale=0.3]{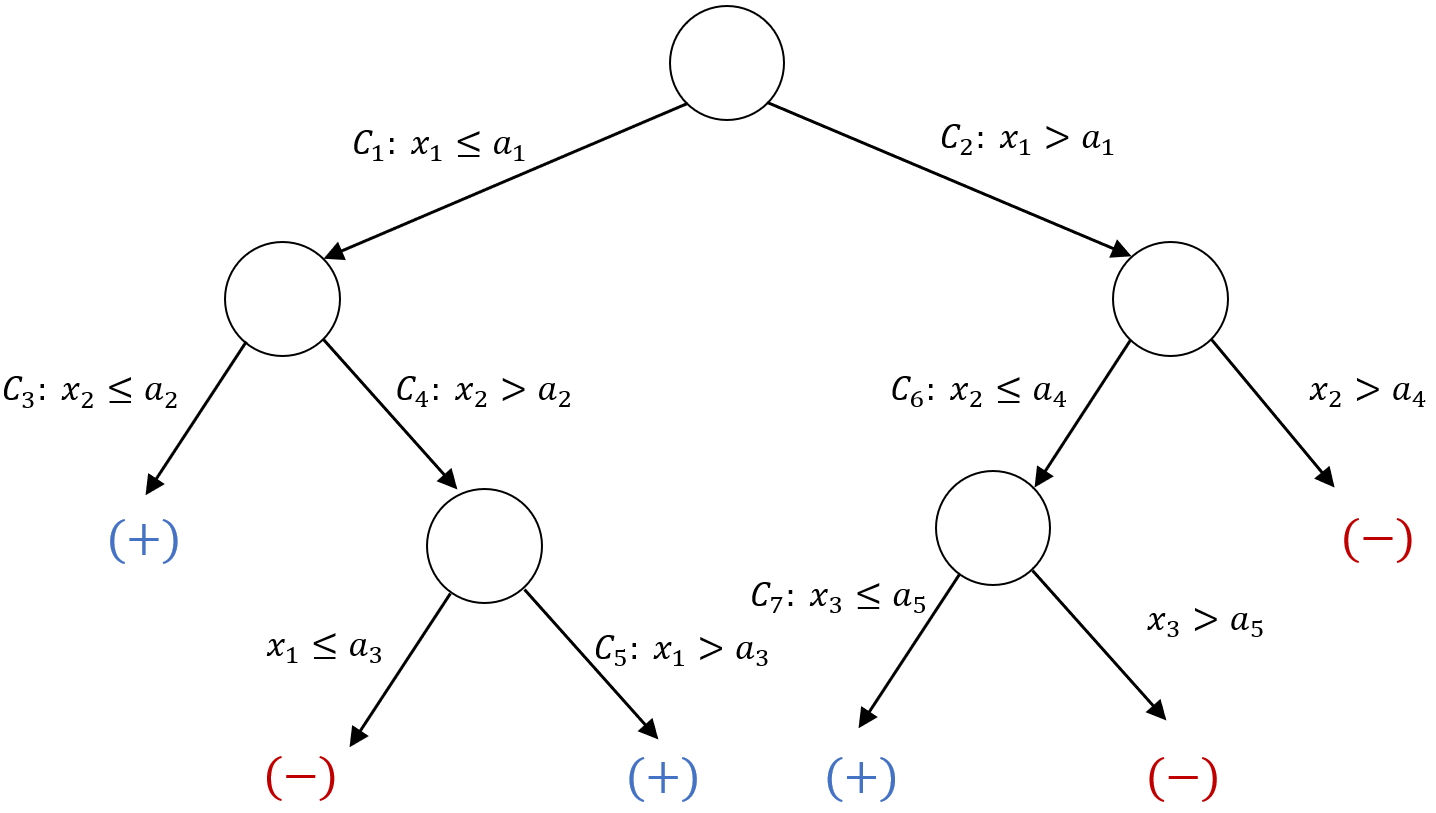}
\caption{Extracting constraints from a decision tree (adapted from \cite{kudla2018one}).}
\label{fig:decTreeEmbed}
\end{figure}

This process is applied to the whole tree until a set of linear constraints capturing the tree is obtained. A similar approach is used in \cite{verwer2017auction} to embed regression trees as constraints. In \cite{thams2017data}, each path from the root to a leaf node is represented by two constraints, with one binary variable per path to determine if a path is chosen. The resulting MILP is solved with Gurobi. \cite{halilbavsic2018data} also follow a similar procedure except that the resulting problem is further reformulated as a Mixed-Integer Second-Order Cone Program (MISOCP) because of the presence of additional non-linear constraints. The resulting MISOCP and MINLP are then solved using commercial solvers.

To embed random forests as constraints, multiple DTs can be used as above, with an additional constraint that selects a consensus based on the outcome chosen by most of the individual decision trees \citep{bonfietti2015embedding}. \cite{biggs2018optimizing} maximize a random forest as the objective. For each tree, they use logical constraints to determine which leaf a solution lies in. For the forest, they use a constraint to ensure that only one leaf for each tree is active. To ensure that the feasible set is bounded and that the solution is not too dissimilar from the observed data, the solution is constrained to be within the convex hull. This is similar to the approach in \cite{mivsic2020optimization} which also has an objective function expressed as the prediction of a tree ensemble, and also uses some form of Benders reformulation with lazy constraint generation to solve the problem. \cite {biggs2018optimizing} also compare solving the MIP for the whole random forest, versus splitting the forest into a number of subsets, and solving each subset using the MIP, and a cross-validation heuristic procedure to achieve performance improvements. In \cite{cremer2018data} where the AdaBoost ensemble learning method is used (with DTs as the weak learners), there is one set of constraints for each DT, and exactly one disjunction must be selected for each learner. Probability estimates for all learners are combined in a weighted sum, and the resulting MILP is solved using a commercial solver. For SVMs, linear or quadratic polynomials can be used directly in the optimization problem, providing the solvers used can handle quadratic constraints.

The advantage of using LP or MILP approaches to learn constraints is seen in this step of the framework, as there are no complex embedding procedures to be followed. In \cite{pawlak2017automatic}, the resolution of the final optimization problem here is relatively easy as the entire constraint learning process was formulated as a MILP from the very beginning. This is also seen in  papers like \cite{sroka2018one} and \cite{schede2019learning}. In \cite{pawlak2021ellipsoidal}, the data is first clustered, and then PCA is used to give each cluster an ellipsoidal shape. These ellipsoids can then be represented using quadratic constraints. Their approach produces Mixed Integer Quadratically Constrained Programs (MIQCP), which can then be solved by any solver that supports quadratic programming.\\

In summary, it can be seen that this step of the constraint learning process is the most complex one. Incorporating predictive models sometimes results in a significant reformulation of the original model. The embedding of the predictive models should however be done in a way that takes full advantage of the optimization solver’s capabilities. This can be done by providing gradients or other useful information directly to the solver or reformulating the problem to become linear, convex, conic quadratic, etc. The approach by \cite{garg2018kernel} results in a quadratic program with linear constraints that is solved using off-the-shelf solvers. Constraint learning approaches that use MILP or LP frameworks such as in \cite{pawlak2017automatic} avoid these complex embedding procedures, however, the disadvantages of this approach could include a lack of flexibility and limited learning ability. There is therefore a trade-off between simplicity and performance to be considered. Wherever possible, it is desirable to keep the final optimization problem in the same complexity class as the original problem.  


\subsection{Verification and Improvement of the Optimization Model}
\label{sec:modelVerifReview}
The validation of the final optimization model is straightforward when dealing with benchmark problems, as the ground truth is already known. This approach to verifying models is seen in a lot of the literature. In optimal power flow, for example, the availability of benchmark problems allows \cite{prat2020learning} to find solutions to problems that were previously intractable. \cite{garg2018kernel} are able to show that their methods obtain near-optimal solutions. The use of benchmark problems also allows for the comparison of the number of synthesized constraints with the actual number of constraints from the benchmark problems \citep{pawlak2017automatic,pawlak2018synthesis,pawlak2019synthesis,karmelita2020cma}. Moreover, they are able to assess the similarity of the syntaxes by computing the mean angle between weight vectors of the corresponding constraints in the synthesized and actual models. 

The feasible regions of the synthesized models can also be compared to those of the actual feasible regions using the Jaccard Index \citep{jaccard1912distribution}, as it is used to measure the similarity between sets. This is used in \cite{kudla2018one}, \cite{pawlak2017automatic} and others. Visual comparisons (limited to 3-dimensions) have also been used to illustrate how well a constraint learning approach captures the real feasible region (e.g. in \cite{pawlak2021ellipsoidal}). In \cite{schede2019learning}, the authors compute the probabilities that feasible (and infeasible) points of the benchmark problem lie in the feasible (and infeasible) region of the learned problem. They also compare the differences between the optimum values of the original and learned problems. \cite{gutierrez2010neural}, \cite{cremer2018data} and \cite{schweidtmann2019deterministic} similarly also test their approaches on benchmark problems, and compare their results to the known solutions of the problems. In addition to using benchmark problems, \cite{xavier2021learning} also evaluate the performance of the method by using out-of-distribution data to measure the robustness against moderate dataset shift.

Using synthetic or well-known benchmark problems allows one to accurately determine how far learned constraints are from the real ones \citep{kumar2019acquiring}, however, this is not always possible. Consequently, other model verification procedures are used in the literature. Simulators can also be used to evaluate constraint learning approaches by comparing learned solutions to solutions from the simulator (e.g. in \cite{verwer2017auction}).

Authors also compare the constraint learning approaches with the typical approaches used to solve those types of problems. This is done in \cite{say2017nonlinear}, for example, where they compare their constraint learning approach with the rule-based approach commonly used to solve their type of problem. \cite{cozad2014learning} implement model improvement by iteratively generating new data and updating the model with these data. The process continues until the error is less than a specified tolerance value. An overview of the approaches used for verification and improvement of the learned models is given in Table \ref{tab:verifUsed}.

\rowcolors{3}{gray!10}{white}
\begin{sidewaystable}
\centering
\caption{Approaches used for model verification and improvement}
\resizebox{\textwidth}{!}{\begin{tabular}{ @{} l *{6}{c} @{} } 
\toprule
  & Benchmark & Compare Constraints & Simulator & Compare to & Adaptive &\\
  & Problems & and/or Feasible Regions & Results & Other Approaches & Sampling & Other\\
\midrule
\cite{bergman2022janos} & x &  &  &  &  & \\
\cite{chen2020input} & x &  & x & x &  & \\
\cite{chi2007modeling} &  &  & x &  &  & \\
\cite{cozad2014learning} &  &  & x &  & x &\\
\cite{cremer2018data} & x &  & x & x &  & \\
\cite{de2003integrating} &  &  & x &  & x & \\
\cite{fahmi2012process} &  &  & x &  &  & \\
\cite{garg2018kernel} & x &  &  & x &  & \\
\cite{grimstad2019relu} & x &  &  &  &  & \\
\cite{gutierrez2010neural} & x &  &  & x &  & \\
\cite{halilbavsic2018data} & x &  &  & x &  & \\
\cite{jalali2019designing} & x &  &  & x &  & \\
\cite{karmelita2020cma} & x & x &  & x &  & \\
\cite{kudla2018one} & x & x &  & x &  & \\
\cite{kumar2019automating} & x &  &  &  &  & \\
\cite{kumar2019acquiring} & x &  &  & x &  & \\
\cite{kumar2021learning} & x &  &  & x &  & \\
\cite{lombardi2017empirical} &  &  & x & x &  & \\
\cite{maragno2021mixedinteger} & x &  &  & x &  & \\
\cite{mivsic2020optimization} &  &  &  & x &  & \\
\cite{venzke2020neural} & x &  &  & x &  & \\
\cite{paulus2021comboptnet} & x &  &  &  &  & \\
\cite{pawlak2017automatic} & x & x &  &  &  & \\
\cite{pawlak2017synthesis} & x & x &  &  &  & \\
\cite{pawlak2018synthesis} & x & x &  &  &  & \\
\cite{pawlak2019synthesis} & x & x &  & x &  & \\
\cite{pawlak2021ellipsoidal} & x & x &  & x &  & \\
\cite{pawlak2021grammatical} & x & x &  & x &  &  \\
\cite{prat2020learning} & x &  &  & x &  & \\
\cite{say2017nonlinear} & x &  &  & x &  & \\
\cite{schede2019learning} & x & x &  & x &  & \\
\cite{schweidtmann2019deterministic} & x &  &  & x &  & \\
\cite{spyros2020decision} & x &  &  & x &  & \\
\cite{sroka2018one} & x & x &  & x &  & \\
\cite{thams2017data} & x &  &  & x &  & \\
\cite{verwer2017auction} &  &  & x &  &  & \\
\cite{xavier2021learning} & x &  &  &  &  & x\\
\cite{yang2021optimization} & x &  &  & x &  & \\
\bottomrule
\end{tabular}}
\label{tab:verifUsed}
\end{sidewaystable}

      \section{Challenges and Opportunities in Constraint Learning}
\label{sec:challengesandOpp}
As the process of constraint learning is relatively new, it stands to reason that there are several challenges or limitations that need to be taken into consideration. The primary challenge is that predictive models are designed for predictive purposes, and not to be embedded into an optimization model. Other challenges that new practitioners should keep in mind include:
\begin{enumerate}
    \item Constraint violation: The optimal solution can be infeasible in reality if the learned constraints do not approximate correctly the true constraints.
    \item Explainability: Although optimization models are usually considered explainable, and therefore understandable to humans, the use of learned constraints makes the optimization model difficult to understand and the optimal solution potentially difficult to explain.
    \item Data availability: A dataset of historical solutions is required to fit a predictive model. Some predictive models require more data than others to have acceptable performances. Obtaining the data for training purposes might be difficult in certain cases. Additionally in certain applications, one-class data (i.e., data on only feasible (or infeasible states)) may be more available than two-class data (i.e., data on both feasible and infeasible states). Potential practitioners of OCL should be aware of what approaches work best in either case.
    \item Computational complexity: Some predictive models require auxiliary variables and many additional constraints in order to be embedded into the optimization model. The number and class of additional constraints and auxiliary variables have a direct effect on the computational complexity of the optimization model.
    \item  Confounders: The use of predictive models to define part of the optimization model should account for potential confounders. It has been noted by \cite{deMast2022} that machine learning models are correlational models and not causal models. The fact that certain values of the features co-occur with certain values of the outcome does not imply that changing the value of the features would change the outcome. These features could be causal, however, there is no guarantee of this. Therefore, whenever a causal model on the features space is available, it should be considered in the embedding of the predictive model.
\end{enumerate}

In light of these challenges, some of the opportunities for further research are presented below. The opportunities are discussed in light of the steps of our proposed framework.\\

\noindent
\textit{Data Gathering and Preprocessing}: There is an opportunity to understand the effects of the data generation and preprocessing phase on the solution of the optimization problem. Is there a particular strategy (e.g. DoCE, Kriging, etc.) that performs best for constraint learning? It is also necessary to understand the size and type of data that is needed in order to get a good predictive model for constraint learning, and not just for prediction.\\

\noindent
\textit{Selection and Training}: Firstly, the choice of which model to use to learn constraints should also be further investigated. Most of the approaches used to learn constraints are the most commonly known methods (regression, ANNs, DTs, etc.), however, there is an opportunity to see how less well-known approaches perform in terms of solution quality, as well as ease of embedding as constraints. Secondly, predictive models often provide a measure of the uncertainty of their predictions. Neural network classifiers, for example, can have Softmax scores \citep{grave17a} while decision trees report the number of misclassified examples at their leaf nodes. Although \cite{cremer2018data} have used probabilistic information from their classifier to good results, further research on how to incorporate such measures of uncertainty into optimization problems should be carried out. Approaches such as robust optimization or stochastic programming might be useful here. Thirdly, as noted above, predictive models are designed for predictive purposes, and not to be embedded into an optimization model. There could be an opportunity to develop models designed to be embedded, as well as models that perform well with small amounts of data. Finally, only a few approaches have used ensemble methods to learn constraints. While not all problems will require the additional predictive power that an ensemble approach brings, it is to be expected that as the best predictive models in ML applications are often ensembles, it makes sense to incorporate their superior abilities into constraint learning.
\\

\noindent
\textit{Resolution of the Optimization Model}: Several approaches for embedding ML models as constraints have been seen, however, can these models be embedded in more efficient ways? The difficulty of using larger neural networks or larger decision trees to learn constraints might be overcome if more efficient embedding procedures are developed. In \cite{schweidtmann2019deterministic}, difficult parts of the neural network ($\tanh$ activation functions) were hidden from the B\&B solver. Further research on hiding unnecessary information from solvers is needed. Further research is also needed on providing additional useful information (such as pre-computed derivatives) to commercial solvers so as to improve solution time and quality. This will be useful when the incorporation of predictive models makes these computations impractical or expensive.\\

\noindent
\textit{Verification and Improvement}: In the reviewed literature, methods for model verification and improvement, usually consist of either comparing to benchmark problems, or to other competing approaches. It could be promising to look into developing formal approaches for verifying learned constraint models. A framework for verifying neural network behavior was developed in \cite{venzke2020verification}, and research could be done to see if similar frameworks can be applied during the constraint learning process. In terms of model improvement, more focus should be put on improving models via processes such as active learning or adaptive sampling. In such cases, it would be interesting to see the effects (if any) of exploration versus exploitation when generating data on the overall solution of the optimization problem. Questions also arise with respect to the robustness of the solution --  What is the sensitivity of the optimal solution with respect to the uncertainty in the learned constraint? How is the robustness of the solution related to the noise in the dataset? A structured and formal approach for verifying optimization problems with learned constraints can help to answer questions such as these.

      \section{Conclusions}
\label{sec:Conclusions}
OCL helps to capture constraints that would otherwise have been difficult to capture \citep{kumar2019automating}, providing there is data available. The benefits are clearly stated by \cite{halilbavsic2018data} where their data-driven approach increases the feasible space, identifies the global optimum, and takes less time than conventional methods. There is however a need to go about the process of OCL in an organized manner, in order to avoid potential pitfalls. We have therefore provided a framework for OCL, which we believe will help to formalize and direct the process of OCL. Besides providing the framework, we reviewed the literature in light of the different steps of the framework, highlighting common trends in constraint learning, as well as possible areas for future research. It is our belief that this paper will help to guide future efforts in OCL, and be of benefit to the wider OR community.

\section*{Acknowledgements}
This work was supported by the Dutch Scientific Council (NWO) grant OCENW.GROOT.2019.015, Optimization for and with Machine Learning (OPTIMAL). The authors would also like to thank Ilker Birbil for his comments during the preparation of the manuscript, and the anonymous referees for useful comments that improved the paper.

    \endgroup

\bibliographystyle{elsarticle-harv} 
\bibliography{References.bib}

\end{document}